\documentclass{article}

 \usepackage[final]{nips_2017}

\usepackage[utf8]{inputenc} 
\usepackage[T1]{fontenc}    
\usepackage{hyperref}       
\usepackage{url}            
\usepackage{booktabs}       
\usepackage{amsfonts}       
\usepackage{nicefrac}       
\usepackage{microtype}      
\usepackage{subcaption}      
\usepackage{multirow,array}
\usepackage{float}
\usepackage{algorithm}
\usepackage[noend]{algpseudocode}

\usepackage{graphicx} 
\usepackage{amsmath}

\newtheorem{defn}{Definition}
\newtheorem{thm}{Theorem}
\newtheorem{cor}{Corrolary}

\usepackage{tikz}
\usetikzlibrary{arrows,positioning}
 \newcommand*\ab{.4}
  \tikzset{
    net node/.style = {circle, minimum width=.5*\ab cm, inner sep=0pt, outer sep=0pt, ball color=black},
    net root node/.style = {net node, minimum width=.5*\ab cm},
    net connect/.style = {line width=1pt, draw=blue!50!cyan!25!black},
  }

\begin{document}

\title{Prosocial learning agents solve generalized Stag Hunts better than selfish ones}  

\author{
Alexander Peysakhovich\\
Facebook AI Research
\And
Adam Lerer \\
Facebook AI Research
}

\maketitle

\begin{abstract}  
Deep reinforcement learning has become an important paradigm for constructing agents that can enter complex multi-agent situations and improve their policies through experience. One commonly used technique is reactive training - applying standard RL methods while treating other agents as a part of the learner's environment. It is known that in general-sum games reactive training can lead groups of agents to converge to inefficient outcomes. We focus on one such class of environments: Stag Hunt games. Here agents either choose a risky cooperative policy (which leads to high payoffs if both choose it but low payoffs to an agent who attempts it alone) or a safe one (which leads to a safe payoff no matter what). We ask how we can change the learning rule of a single agent to improve its outcomes in Stag Hunts that include other reactive learners. We extend existing work on reward-shaping in multi-agent reinforcement learning and show that that making a single agent prosocial, that is, making them care about the rewards of their partners can increase the probability that groups converge to good outcomes. Thus, even if we control a single agent in a group making that agent prosocial can increase our agent's long-run payoff. We show experimentally that this result carries over to a variety of more complex environments with Stag Hunt-like dynamics including ones where agents must learn from raw input pixels.
\end{abstract}

\section{Introduction}
Constructing agents which can make good decisions in complex environments is a cornerstone of modern artificial intelligence research. An increasingly popular paradigm for this task is deep reinforcement learning (deep RL, \cite{kaelbling1996reinforcement,sutton1998reinforcement,silver2016mastering,mnih2015human}). Recent work has become interested in using deep RL to construct agents that can function well in environments which include other agents \citep{silver2016mastering,lazaridou2017multi,das2017learning,lowe2017multi,foerster2017counterfactual,evtimova2017emergent,foerster2016learning,havrylov2017emergence,lewis2017deal,heinrich2016deep,peng2017multiagent,leibo2017multi}. In many of these examples deep RL been applied by the method of reactive training where the learner treats other agents as part of the environment and optimizes its own reward. Reactive training works very well in zero-sum situations, where for one agent to gain the other must lose \citep{silver2016mastering,wu2016training,neumann1928theorie}. However, for general-sum games good outcomes are far from guaranteed \citep{sandholm1996multiagent,leibo2017multi,lerer2017maintaining,foerster2017learning}. In this paper we will study a simple modification to independent learning that can be applied to a single agent out of a dyad and can lead to better outcomes in a class of coordination games.

As with the works above we are interested in situations where we construct an agent that will go into an initially poorly understood environment (which we refer to as a game). Our agent must learn from its experiences to update its policy and maximize some scalar reward. However, there will also be other agents which we do not control and will also learn from their experiences using a form of reactive learning. Our starting point will be standard independent multi-agent reinforcement learning (independent MARL). In independent MARL each agent treats the other agents simply as a fixed part of the environment.\footnote{Standard RL techniques assume that the underlying environment is stationary and so MARL can have convergence issues due to non-stationarity of the learning environment \citep{lanctot2017unified,lowe2017multi,foerster2017counterfactual}. These algorithmic issues are very important but somewhat orthogonal to the game theoretic issues we discuss in this paper.}

If all agents use independent MARL, and if the system converges to deterministic strategies, then it will have converged to a Nash equilibrium of the game \citep{fudenberg1998theory} - a set of strategies such that no player has incentive to unilaterally deviate.\footnote{Note that computing a Nash equilibrium is in general computationally difficult and so we cannot guarantee convergence for all games in a reasonable time \citep{daskalakis2009complexity}.} In general-sum games there can be multiple equilibria and thus initial conditions can alter the long-run payoffs that our agent (and others involved) can earn. In addition, in some classes of games many different kinds of dynamics can converge on equilibria with low payoffs even though high payoff equilibria exist \citep{kandori1993learning,fudenberg1998theory}. We will focus on one such example: Stag Hunt games.

In the simple, matrix-form, two-player Stag Hunt each player makes a choice between a risky action (hunt the stag) and a safe action (forage for mushrooms). Foraging for mushrooms always yields a safe payoff while hunting yields a high payoff if the other player also hunts but a very low payoff if one shows up to hunt alone. There are two Nash equilibria: either both players show up to hunt (this is called the payoff dominant equilibrium) or both players stay home and forage (this is called the risk-dominant equilibrium \cite{harsanyi1988general}). 

In the Stag Hunt, when the payoff to hunting alone is sufficiently low, dyads of independent learners as well as evolving populations converge to the risk-dominant (safe) equilibrium \citep{kandori1993learning,nowak2006evolutionary,fudenberg1998theory,matignon2012independent}. The intuition comes from the fact that even a slight amount of doubt about whether one's partner will show up causes an agent to choose the safe action. This in turn causes partners to be less likely to hunt in the future and the system trends to the inefficient equilibrium.

Failure to coordinate on good outcomes in the Stag Hunt (and related games) has been extensively studied outside of MARL in many fields, particularly in the social and behavioral sciences \citep{kandori1993learning,carlsson1993global,van1990tacit,camerer2003behavioral,yoshida2008game} because the Stag Hunt is a simple example of a more general type of coordination dynamic which is ubiquitous in real world interactions. For this reason how agents can avoid the spiral to the inefficient equilibrium is a problem of interest across many fields. 

Existing work in RL has proposed modifications to independent MARL to increase the probability of the system converging to a preferred equilibrium \citep{littman2001friend,babes2008social,devlin2011theoretical,devlin2011empirical,kapetanakis2002reinforcement,yoshida2008game,panait2008theoretical,matignon2012independent,wang2003reinforcement}. One prominent set of work has investigated endowing agents with a form of optimism, eg. lenient learning \citep{panait2008theoretical,matignon2012independent} or Frequency Maximum Q-learning \citep{kapetanakis2002reinforcement}, while another has focused on potential-based reward shaping \citep{babes2008social,devlin2011theoretical,devlin2011empirical}. We will be interested in building upon ideas from the literature on reward shaping and applying these insights to situations beyond those typically considered. In particular, we are interested in environments where we only control a single agent, where decisions are not single actions but rather temporally extended sequences, and where good policies may need to be learned using function approximation (eg. directly from raw pixels).

Reward shaping methods change the updating rule of all agents in the system using a potential function which encodes heuristic knowledge about `good' states. If we know the structure of the game, we can choose an appropriate potential function \citep{babes2008social,devlin2011theoretical,devlin2011empirical}. A special property of Stag Hunt games is that changing both agents to optimize the joint reward changes the basin of attraction of the payoff-dominant equilibrium \cite{yoshida2008game}. We can restate this idea as making agents prosocial: that is, changing their utility function to care not only about their reward but also their partner's rewards .

Our first contribution is an extension of existing results to the case where we control a single agent. We show analytically that in $2 \times 2$ Stag Hunts we can improve the probability of convergence to a good equilibrium even if we can only make a single agent prosocial. We extend this result to $N$ strategy $2$ player games as long as any $2 \times 2$ sub-game is a Stag Hunt. This occurs because in Stag Hunts equilibria can be ranked by their payoffs and both agents agree on this ranking, thus, causing one agent to `learn towards' one equilibrium causes others to follow.

We then begin to probe some of the limitations of the analytical result experimentally. In particular, we ask what happens when Stag Hunts are played by more than $2$ agents. We focus on two studied extension of the Stag Hunt: games on networks \citep{jackson2010social,morris2000contagion,chen2009efficient} and multi-player `weak link' games \citep{van1990tacit,camerer2003behavioral,van1991strategic}. 

Our second contribution is to experimentally generalize these insights in a domain where analytical solutions are difficult: Markov games with Stag Hunt-like structure and learning via function approximation. We consider a set of two-player grid based games as well as an Atari game and show that in each of these cases adding prosociality helps policy gradient to converge to good outcomes. Each group of agents either converges to always playing the the safe or risky equilibrium, so improving the probability of converging to a good outcome also improves the long run payoff of the prosocial agent. Thus even in complex Stag Hunt-like games, even if we control just a single agent, and \textit{even if we only care about that agent's payoff} it can still be better to make that agent prosocial in the long-run.
 
\section{Risk Dominance, the Stag Hunt, and Equilibrium Selection}
We begin by illustrating the intuition behind our results with the simple matrix Stag Hunt. In the Stag Hunt players simultaneously decide to either take a risky option (Hunt the Stag) or a safe option (Forage). Foraging always yields a safe payoff, whereas Hunting yields a higher payoff if the other person also hunts but yields a bad payoff if the player shows up to hunt alone (because the Stag gores her). This is illustrated in the payoff matrix below (rows represent strategy choices for player $1$, columns represent strategy choices for player $2$, entries represent payoffs to each player as a function of strategies chosen):

  \begin{table}[H]
  \begin{center}
    \setlength{\extrarowheight}{2pt}
    \begin{tabular}{*{4}{c|}}
      \multicolumn{2}{c}{} & \multicolumn{2}{c}{Player $2$}\\\cline{3-4}
      \multicolumn{1}{c}{} &  & $Hunt$  & $Forage$ \\\cline{2-4}
      \multirow{2}*{Player $1$}  & $Hunt$ & $(h,h)$ & $(g,c)$ \\\cline{2-4}
      & $Forage$ & $(c,g)$ & $(m,m)$ \\\cline{2-4}
    \end{tabular}
   \end{center}
  \end{table}
  
\begin{defn} 
A $2 \times 2$ game is a \textbf{generalized Stag Hunt} if $h > c \geq m > g$.
\end{defn}
  
Let $A_1, A_2$ be the action spaces of the players and let $R_i (a_1, a_2)$ be the reward players receive from a pair of actions. We consider the set of stable points in this game:

\begin{defn}
Nash equilibria are strategy pairs $(a^*_1, a^*_2)$ such that for any $a'_1$ we have $$R_1 (a^*_1, a^*_2) \geq R_1 (a'_1, a^*_2)$$ and for any $a'_2$ we have $$R_2 (a^*_1, a^*_2) \geq R_2 (a^*_1, a'_2).$$ 
\end{defn}

In a Nash equilibrium, neither agent has incentive to deviate from their chosen strategy given what the other agent is doing. For this reason, they are possible convergence points of a learning process. There are two (non-randomized) equilibria here: both choose to Hunt or both choose Forage. Hunting yields higher payoff for both agents (is the \textit{payoff-dominant} equilibrium) thus one may hope that learning agents converge to Hunt. However this may not always be true. 

The generalized Stag Hunt conditions imply two things. First, that $(Hunt,Hunt), (Forage,Forage)$ are the equilibria and $(Hunt,Hunt)$ is payoff dominant. The second part of the condition $b \geq d$ implies that the coordination game has a sort of asymmetric risk, if one player chooses $Hunt$ but the other chooses $Forage$, only the $Hunt$ player loses. Thus $Forage$ is a safe action that does well no matter what the other chooses whereas $Hunt$ is a risky action that only works if the other person also hunts. 

We can now consider which equilibrium is more likely to be selected by a general class of learning dynamics. This example captures much of the intuition behind existing results about why dynamics often converge to safe outcomes \citep{kandori1993learning,nowak2006evolutionary,fudenberg1998theory} as well as the basic idea behind why certain modifications to independent MARL (eg. \cite{devlin2011theoretical,devlin2011empirical,kapetanakis2002reinforcement,yoshida2008game,panait2008theoretical,matignon2012independent}) can help drive the system towards payoff dominant equilibria. 

We assume that agents play the game repeatedly (either with each other, themselves in a `self-play' form or with a population of others).\footnote{This setup does not allow agents to condition their behavior on histories of play (except through the learning rule). In general, allowing such conditioning greatly increases the strategy space and can create new equilibria \citep{fudenberg1986folk}. For example, in social dilemmas (eg. the repeated Prisoner's Dilemma) allowing conditioning on histories can change the set of attainable payoffs. This change occurs because the socially optimal outcome of mutual cooperation is not an equilibrium in the one-shot game but can be supported in the repeated game. However, in Stag Hunts allowing history-conditional strategies does not create `better' equilibria as the best possible outcome (joint hunting) is already an equilibrium in the one-shot game. For this reason we will follow the literature and used the standard setup of `games that are played repeatedly without being repeated games' \citep{fudenberg1998theory}.} Agents start with beliefs $p_1, p_2$ which are the probabilities they expect their partner to choose $Hunt$. Agents best respond to this belief. Having observed the choice of their partner, they update their beliefs in the correct direction (that is, if the partner chose $Hunt$, $p$ goes up). This means that agents choose $Hunt$ if $p_i h + (1-p_i) g \geq p_i c + (1-p_i) m.$ We can calculate the critical value of $p^*$ such that if $p > p^*$ the agent chooses $Hunt$ otherwise they choose $Forage$.\footnote{Note that we ignore mixed equilibria in our analysis as they are extremely unstable: for a player to play a mixed strategy they must be indifferent between all strategies in the support and this set of beliefs is highly unstable in the Stag Hunt where learning dynamics push both players towards pure strategy equilibria.}

\begin{figure}[h]
\centering
\includegraphics[scale=.35]{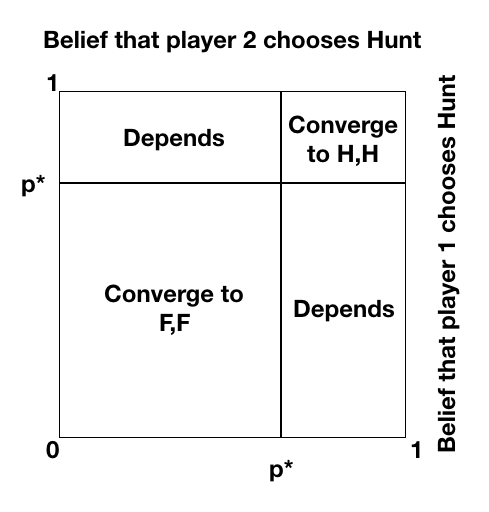}
\caption{Belief-based learning dynamics in the Stag Hunt game depend on initial states as well as all entries of the payoff matrix, this same intuition applies to RL-based dynamics which simply use values of actions.}
\label{stag_basins}
\end{figure}

This tells us something about the basins of attraction of each equilibrium. If both agents have beliefs above $p^*$ then they will converge to $(Hunt,Hunt)$ - this is because they will play $Hunt$ this period and only increase their beliefs and thus they will play $Hunt$ next period and so on. A similar argument holds if both agents start below $p^*$. If one agent is above and the other is below then the specifics of the learning rule come into play and limit behavior depends on which of the two attractor regions they reach first. Note that what matters for the computation of $p^*$ are not just $h$ and $m$ but also the off-equilibrium payoffs $g$ and $c$.\footnote{A particularly important related concept is risk dominance \citep{harsanyi1988general} which asks whether $p^* > .5$ (in which case $Hunt$ is said to be risk-dominant) or not (in which case $Forage$ is risk dominant). Work in evolutionary game theory shows that in large population evolutionary processes in $2 \times 2$ games with multiple equilibria it is risk dominance not payoff dominance that serves as the unique convergence point \citep{kandori1993learning,nowak2006evolutionary,fudenberg1998theory}. These results typically focus specifically on low mutation rates and large populations, however their qualitative results carry over to many other learning dynamics. Indeed, another way to understand risk dominance is to think about it as the basin of attraction of each equilibrium. The risk dominant equilibrium in a symmetric game is the one with the strictly larger basin of attraction.} This is illustrated in figure \ref{stag_basins}. Interestingly, in games with more than two strategies the basins of attraction of various equilibria can be informed even by the presence of strategies which are never played in any equilibrium as well as the exact properties of the dynamics in question \citep{panageas2016average,nowak2006evolutionary,kandori1993learning}.

With this notion of basin of attraction in mind, we ask whether modifying our agents while keeping the game constant can shift the basin of attraction. We consider a type of agent with social preferences:

\begin{defn}
A prosocial agent's total utility from an outcome is given by $$U_i (a_i, a_j) = (1-\alpha) R_i (a_i, a_j) + \alpha R_j (a_i, a_j).$$ We refer to $\alpha$ as the agent's level of prosociality. $\alpha = 0$ agents are perfectly selfish, $\alpha = .5$ agents are fully prosocial (weigh their and their partner's utility equally) and $\alpha = 1$ agents are selfless.
\end{defn}

We now show our main analytical result as well as a corollary which motivate our experiments. We relegate the proofs to the Appendix.

\begin{thm}
In a generalized Stag Hunt the size of the basin of attraction of $(Hunt,Hunt)$ increases in both players' level of prosociality. There exists $\bar{\alpha} \in (0, 1]$ such that if either agent has $\alpha \geq \bar{\alpha}$ the unique interior attractor is $(Hunt,Hunt).$
\end{thm}

\begin{cor}
In any symmetric $N\times N$ game where the only pure equilibria are symmetric and the payoff matrix subset to any pair of strategies is a generalized Stag Hunt there exists $\bar{\alpha} \in (0,1]$ such that if for either player $\alpha > \bar{\alpha}$ the unique convergence point is the payoff dominant equilibrium.
\end{cor}

Theorem 1 and its Corollary implies that in generalized Stag Hunts, any agent can increase their (expected) payoff after convergence by increasing their $\alpha$. However, in games with more complex strategy spaces where some aspects of the game do not obey the Stag Hunt inequality, increasing $\alpha$ may be counterproductive. We discuss this in more detail in the conclusion. When $\alpha$ is close to $1$ the agent may learn a policy that is both bad for itself and socially inefficient. Considering $\alpha \le 0.5$ at least guarantees that the agent will only choose to hurt its own payoffs if the resulting outcome is efficient (increases the sum of the payoffs). Thus we mostly think about prosocial, but not selfless, agents in practice and look at $\alpha \in \lbrace 0, .5 \rbrace$ in our experiments. 

\section{Experiments}
\subsection{Matrix Stag Hunt}
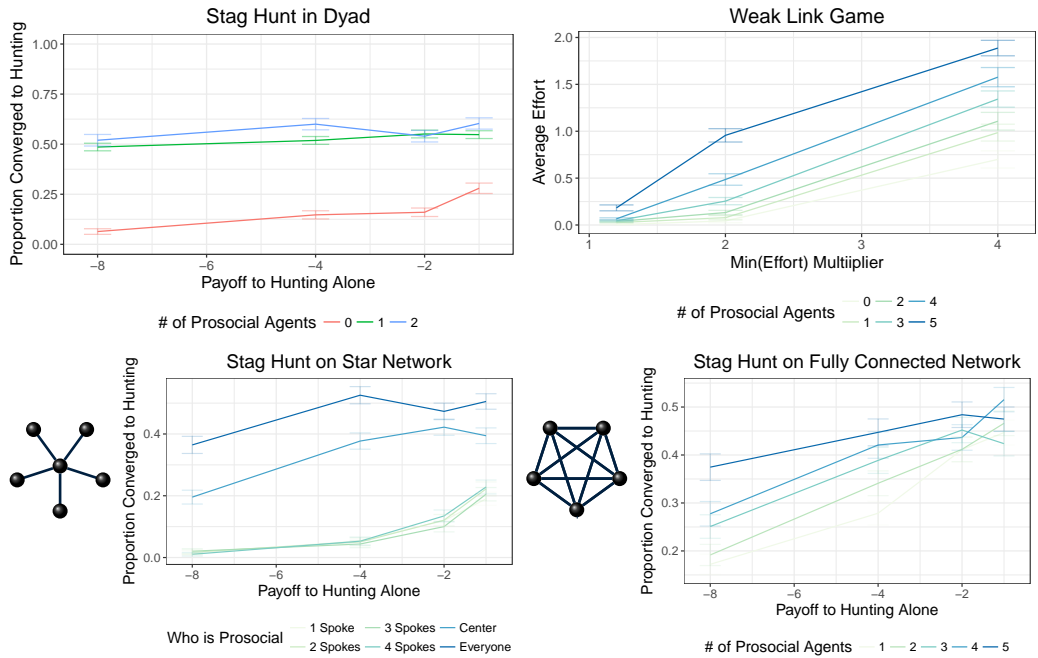
\begin{figure*}[ht!]
\centering
\includegraphics[scale=.45]{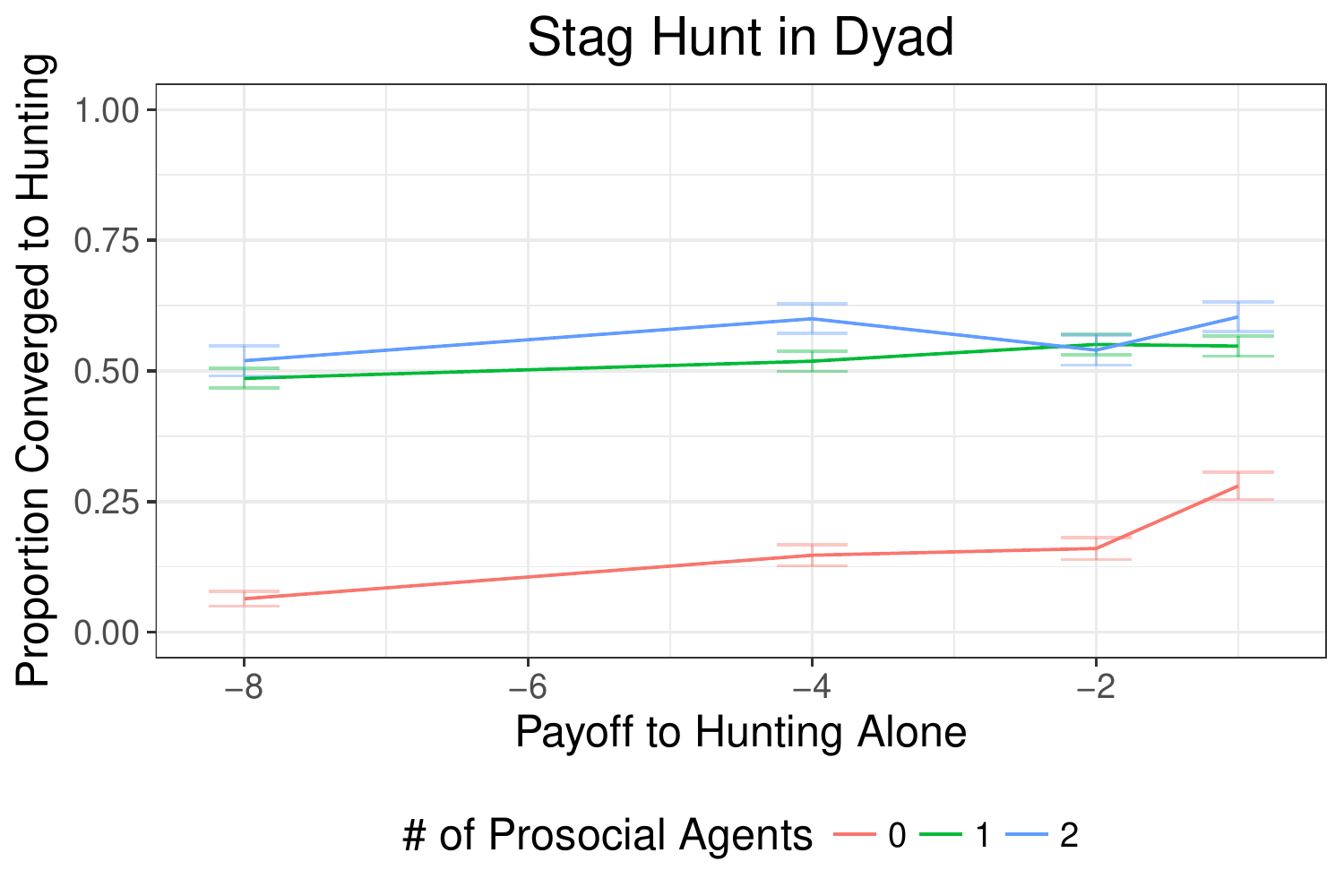}
\includegraphics[scale=.45]{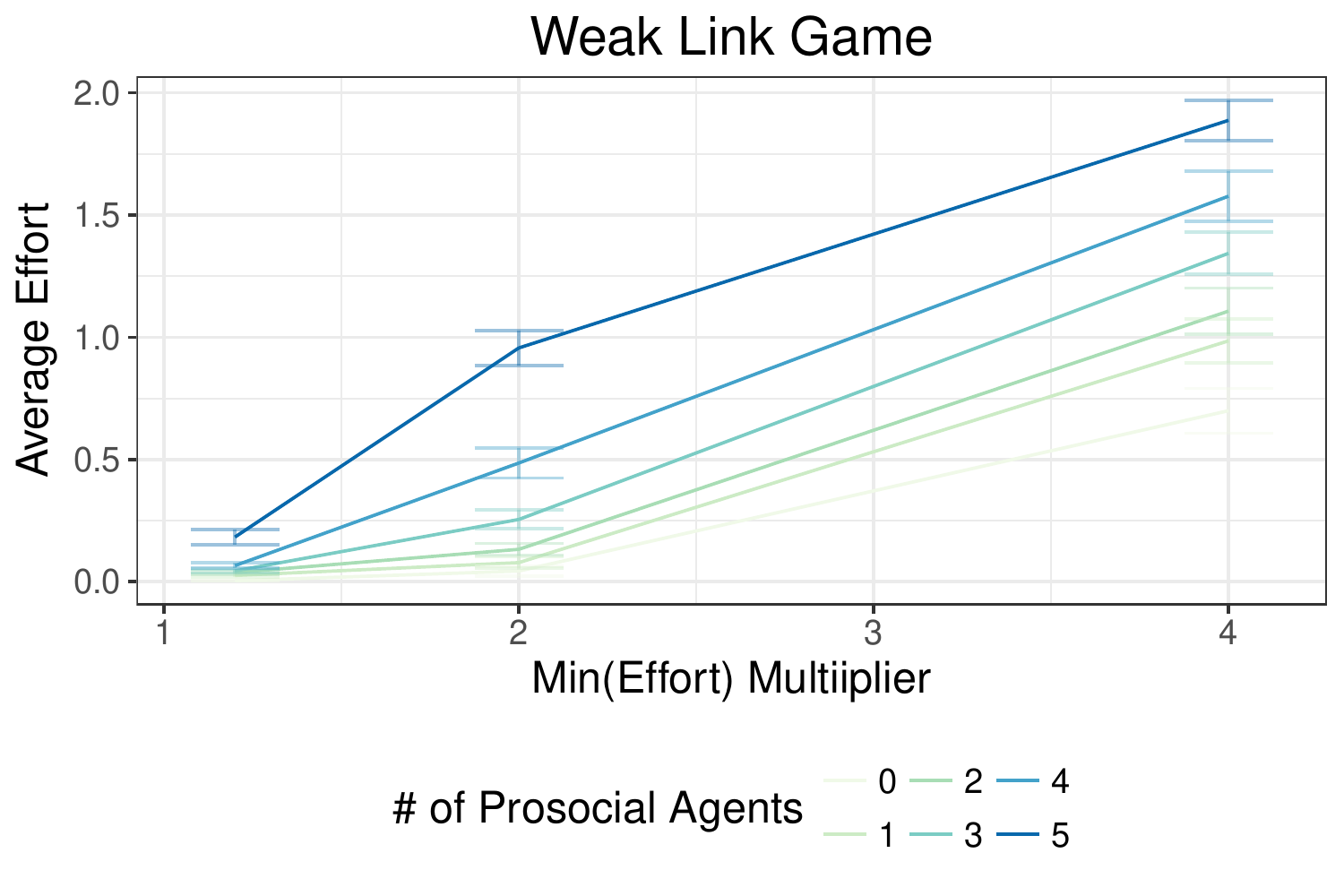} \\
\raisebox{2cm}{\begin{tikzpicture}[scale=0.3]
      \node (root) [net root node] {};
      \foreach \i in {0,...,4}
        \path [net connect] (root) -- (-90+\i*72:2) node [net node] {};
    \end{tikzpicture}}
	\includegraphics[scale=.42]{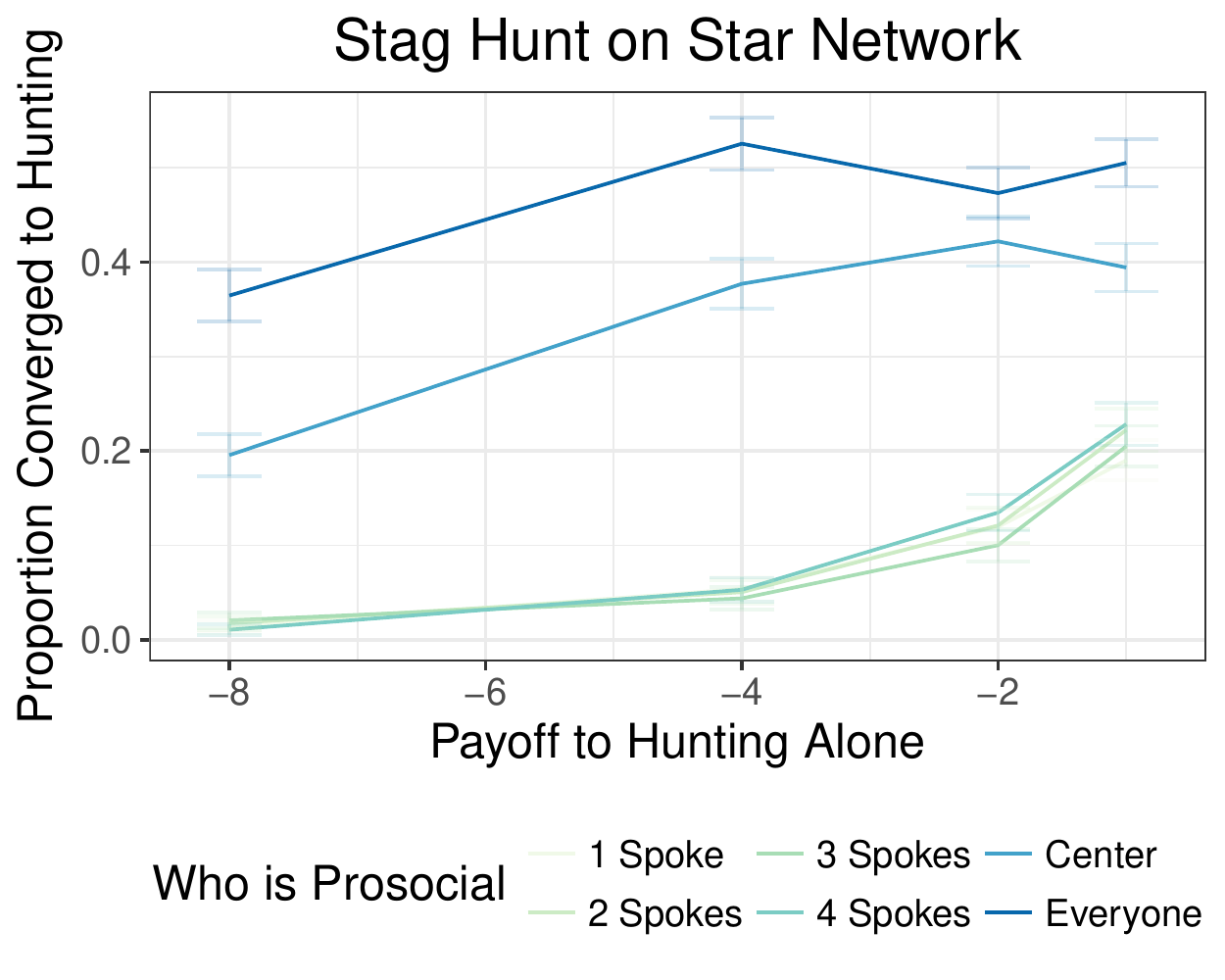} 
	\raisebox{2cm}{\begin{tikzpicture}[scale=.3]
      \foreach \i in {0,...,4}
        \path (-90+\i*72:2) node (n\i) [net node] {};
      \foreach \i in {0,...,4}
        \foreach \j in {0,...,4}
          \path [net connect]
            (n\i) -- (n\j);;
    \end{tikzpicture}}
        \includegraphics[scale=.42]{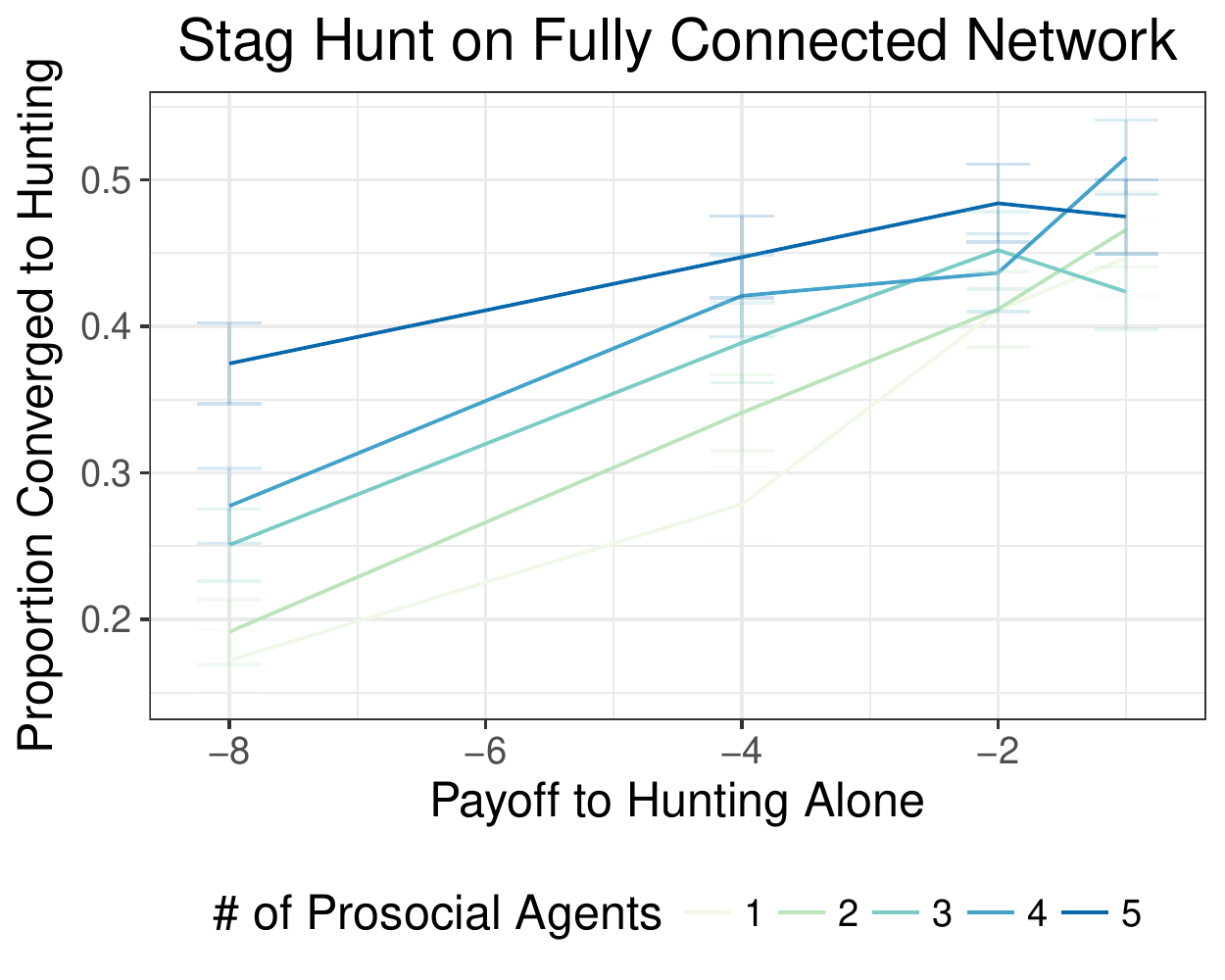}
\caption{In the Stag Hunt, selfish agents trained reactively with policy gradients fail to do well but the addition of social preferences to even one of the agents helps guide convergence to a good outcome though does not guarantee it. Expanding to Stag Hunts beyond dyads shows us that sometimes prosociality of a single agent can greatly affect outcomes, such as in the case of the central agent on a star network, but other times even many prosocial agents may not help, such as in the weak-link game or in the fully connected network Stag Hunt.}
\label{staghunt_results}
\end{figure*}

Our theorem implies that if we endow agents (and more importantly, even one agent) with prosocial preferences then dynamics should now more reliably select the payoff dominant equilibria in $2 \times 2$ matrix games. The relative improvement depends on the learning rule used and details of the payoff matrix, so we begin by studying policy gradient in the $2 \times 2$ Stag Hunt.  We set $h$ (the payoff from joint hunting) to $2$, $m$ and $c$ (the payoff from foraging regardless of what the other person is doing) to $1$ and vary the $g$ (hunting alone and getting gored by the stag) payoff as well as the prosociality of both players. We train policy gradient based agents with softmax action implementation. For optimization we use Adam \citep{kingma2014adam} with a learning rate of $.01$ as well as random initialization. We train each pair of agents for $400$ rounds and look at average behavior in the last $50$ rounds. We train $300$ replicates and average them together.

We see from figure \ref{staghunt_results} (left) that both $g$ and social preferences affect convergence to the payoff dominant equilibrium. There are situations where we control both agents (eg. in using self play to find good policies). In this case, we see that adding social preferences to both agents can help find good optimal equilibria but actually in the dyadic case having a single agent with prosocial preferences is almost as efficient as having both be prosocial. 

\subsection{Prosociality With More than 2 Players}
The results above show that a single prosocial agent can help in the matrix, dyadic Stag Hunts. An important question is how robust these results are to changes in the game complexity as well as the number of players. We move to Stag Hunts with more than $2$ agents played on networks. 

There is a large literature about coordination games on networks that looks at questions including which agents should be `seeded' with an initial behavior to get it to spread on a network  \citep{jackson2010social, morris2000contagion,chen2009efficient}. A major takeaway from this literature is that seeding more central agents leads to larger cascades or higher probability of diffusion. 

We ask whether a similar relationship holds for the question of making subsets of agents prosocial. We consider the standard coordination games on graphs setup \citep{jackson2010social}: agents live on a graph, in each round each agent chooses an action from $Hunt$ or $Forage$ and plays the Stag Hunt with all agents that it is connected to using that single action. The agent receives the average (or total) payoff from these games and updates its strategy for the next iteration. We use the same algorithm, parameters, and analysis as the dyad Stag Hunt above. We consider $5$ agents playing on two graphs: either a fully connected graph or a star network. We vary the loss to hunting alone as well as how many (and which) agents we make prosocial. 

We also consider the weak-link game which is a type of multiplayer Stag Hunt \citep{van1990tacit,camerer2003behavioral,van1991strategic}. Here there are $5$ players. Each simultaneously chooses an effort level $e_i \in \lbrace 0, \dots, 5 \rbrace.$ Players pay a linear cost of effort. This effort goes into a project and the project's success is determined by the weakest link, that is the player who put in the least effort. 
\begin{figure*}[ht!]
\centering
\includegraphics[scale=.45]{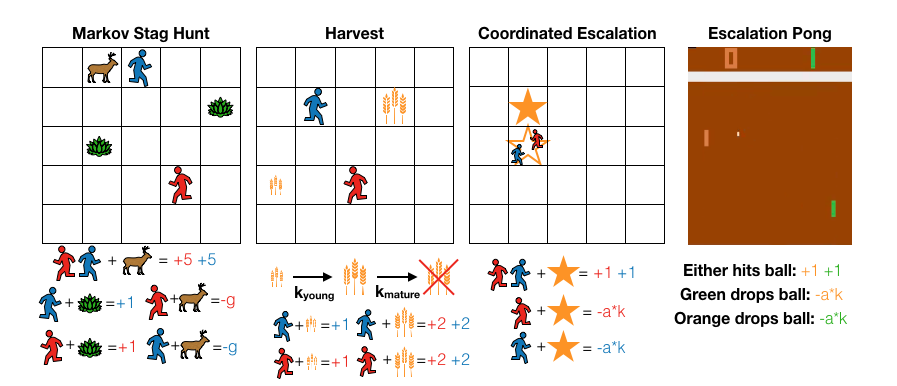}
\caption{Markov games that have complicated policy spaces but preserve high level properties of the Stag Hunt. In each game agents have access to safe policies (which yield low rewards but work no matter what the other agent does) or high reward policies that require coordination with the other agent and can lead to very poor payoffs if the coordination fails.}
\label{markov_games}
\end{figure*}
At the end of the round each agent receives a payoff given by $A \min_{j} (e_j) - e_i.$ This game has Stag Hunt properties because the best equilibrium is one where all agents choose to put in maximum effort but any identical effort levels form equilibria. We apply policy gradient with $5$ players and vary $A$ (which can be thought of as highly related to the risk parameter) as well as agent prosociality. Note that here we have a choice of how to do prosocial payoffs: we can either combine the agent's reward with the average of everyone's reward (which corresponds to the agent weighing their own utility against some total `global utility') or use the sum of everyone's reward (which would cause the prosocial agent to be fully utilitarian but with a large number of other agents would cause the agent to ignore their own reward for the common good). For the reported experiments we use the average and run $100$ replicates.

The games and the results are shown in Figure \ref{staghunt_results}. We see that in the case of the star network making just making the center agent prosocial can greatly increase the probability that the whole group converges to the payoff dominant outcome. However, in the fully connected network the power that an individual agent has over where the group converges is quite limited. In the weak-link game even making all agents prosocial and making the payoff multiplier high still results in poor coordination on the best equilibrium. This shows that in more complex situations many aspects of the underlying game can greatly change the ability of a single (or few) prosocial agents to affect outcomes.

\begin{figure*}[ht!]
\centering
\includegraphics[scale=.45]{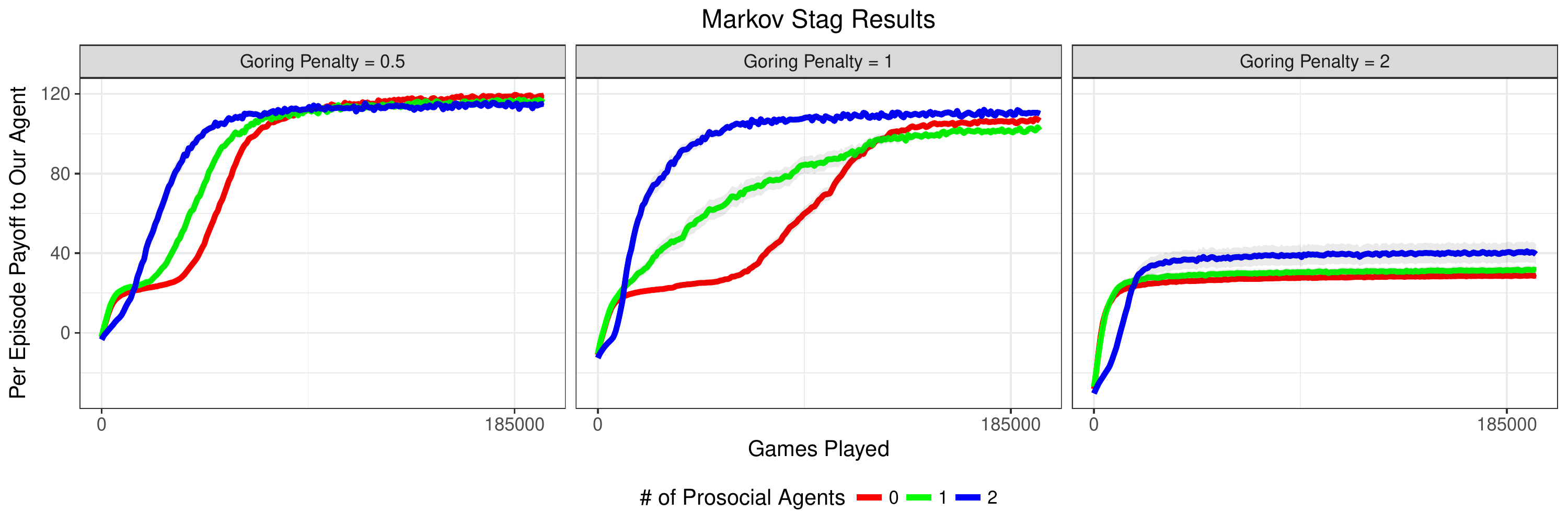}
\includegraphics[scale=.45]{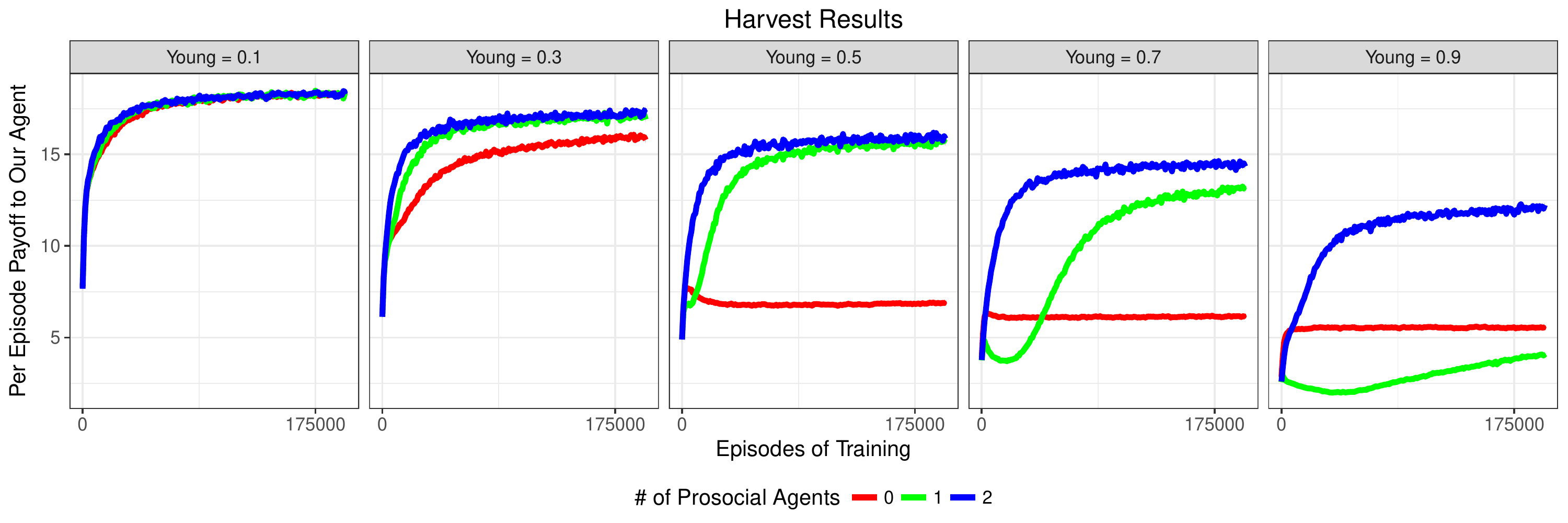}
\includegraphics[scale=.45]{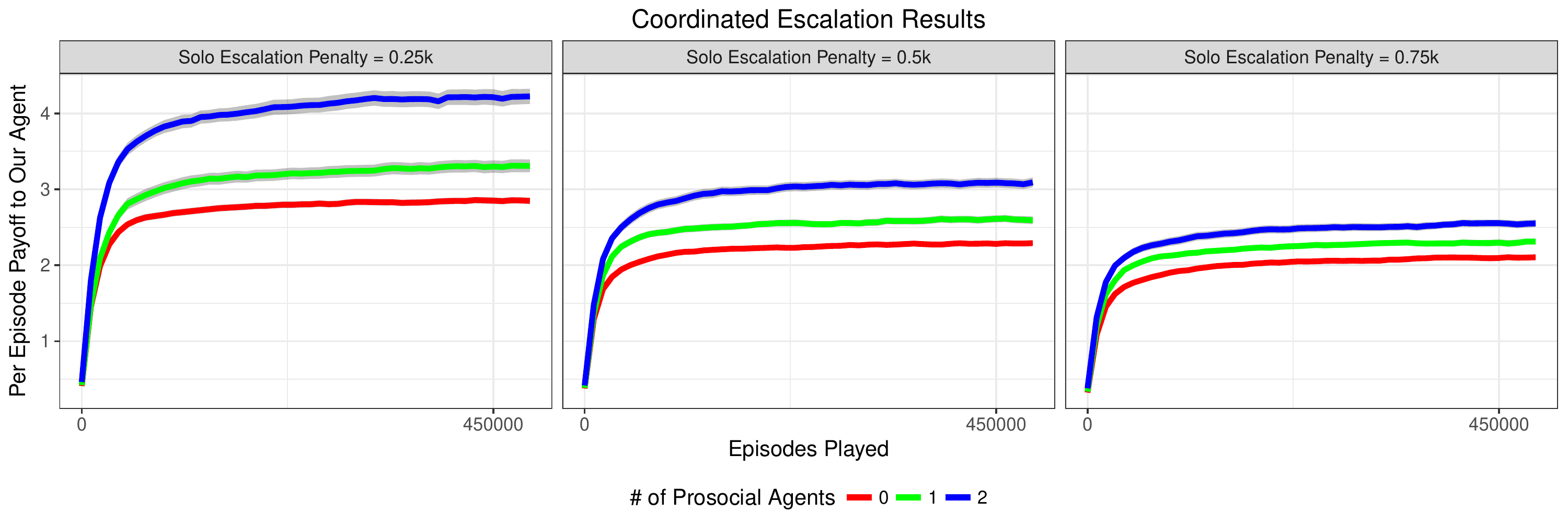}
\includegraphics[scale=.45]{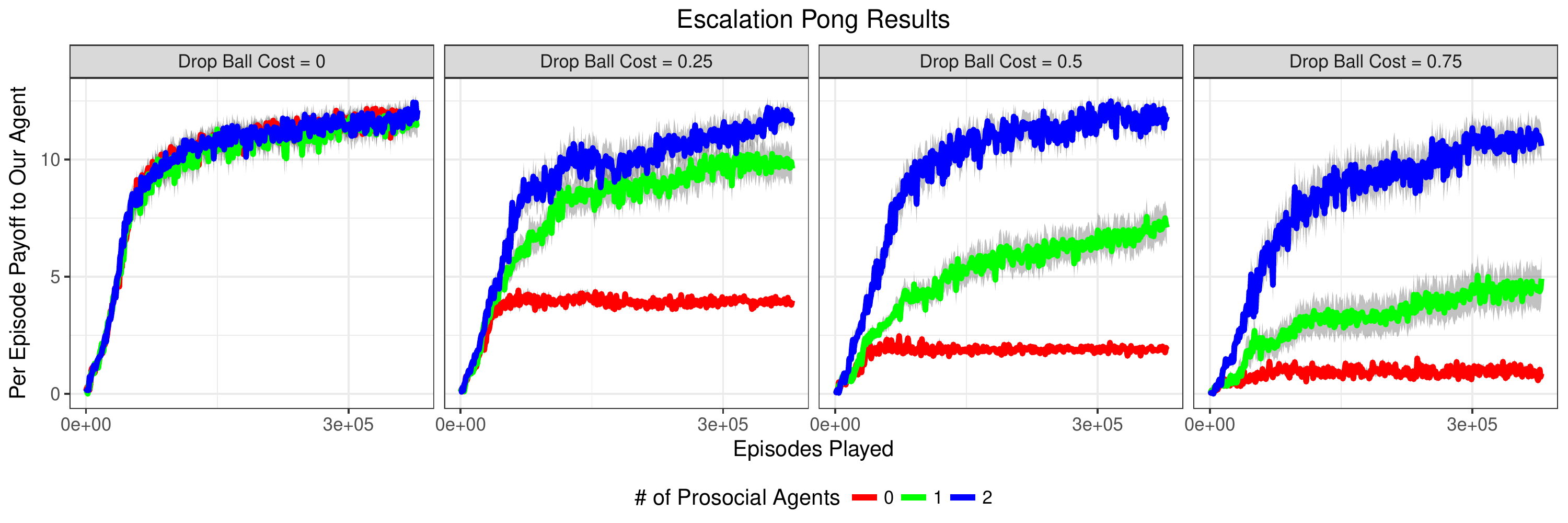}

\caption{We see that the general pattern from the $2 \times 2$ Stag Hunt generalizes to the Markov games. Increases in risk lead to convergence to equilibria with less coordination while giving prosocial preferences to agents increases the dyad's ability to coordinate on good outcomes. The weakest effect exists in our Markov Stag Hunt as well as Harvest with extremely high risk. Lines reflect average payoffs over replicates smoothed over $1000$ episode blocks. Error bars reflect standard errors estimated using independent replicates.}
\label{harvest_results}
\end{figure*}

\subsection{Markov Games with Stag Hunt Properties}

We now consider a set of more complex two-player Stag Hunt-like games. Each of these games we model using the framework of Markov games. In each of these cases we apply a policy gradient with state representation constructed using a deep convolutional network with inputs as the raw board pixels. Because we use standard setups from the literature we put the details of the model architecture/training in the Appendix. 

We consider $3$ different grid-world games: Markov Stag Hunt, Harvest and Escalation (Figure \ref{markov_games}). In each of these games two agents move on a $5 \times 5$ grid and can move in any of the $4$ cardinal directions. In the Markov Stag Hunt the grid is populated with a Stag and $2$ plants. Moving over a plant gives either agent $1$ point and causes the plant to disappear and re-appear in another part of the board. Moving over the Stag causes an agent to lose $g$ points but if both agents move over the stag simultaneously they each gain $5$ points and the stag disappears and re-appears randomly elsewhere on the board. In each time period the stag moves towards the closest agent to it, although the stag can never catch an agent who continues to move away from it. 

Like in the standard Stag Hunt, there are two equilibria here - either both agents try to get the stag or they both try to stay away from it and pick up plants. There is asymmetric risk, if one chooses to try to hunt while the other avoids, the hunter will suffer the consequences.

In the Harvest game at each time step a plant can appear randomly somewhere on the board (up to $4$ plants can be on the board at a time). Each plant is born `young', then every time step turns `mature' with probability $r_{mature}$. While a plant is mature it can die on each time step with probability $r_{death}$. The probabilities are always selected such that each plant lives for $20$ time steps in expectation.

Players can move over plants to pick them up. Players receive $1$ point if they up a young plant, however waiting until each plant becomes mature and picking it up yields $2$ points to \textit{both} players. Thus the coordination question is whether to wait or rush for the plants. Choosing to wait is an equilibrium that yields high payoffs for both players. However, just like stag hunting it incurs a risk that one's partner loses their nerve and grabs the young plant. 

In the Coordinated Escalation game a special marker appears on one of the squares. If the agents step on the square together, they both receive one point, at which point an adjacent square lights up. If the agents step together onto the next square, they receive $1$ point. If at any time an agent breaks the streak (eg. by stepping off the path), the other agent receives a penalty of some multiplier times the current length $T$ of the streak and the game ends. The current streak length $T$ is observed (encoded in the state). This game has many equilibria, where both agents play to keep streaks of size $T$ but no more with risk escalating at each time step (as the reward from further escalation is always $1$ but the cost from one's partner failing to continue the pattern increases linearly). 

We also consider a version of Escalation where agents must learn from raw pixels. We use methods employed in \cite{tampuu2017multiagent} to adapt Atari Pong to construct Escalation Pong. In Escalation Pong each agent controls a player, each time the ball is hit both agents receives a reward of $1$, however if an agent drops the ball (allows it to pass) then \textit{the other agent} receives a reward of $-k$ where $k$ is proportional to the number of times the ball has been hit back and forth. The game can end stochastically at any time period and also ends when any player drops the ball.

We again compare the performance (here in terms of payoff to our agent) from situations where both agents are selfish, both agents are prosocial, and where only our agent is prosocial. In all conditions, both agents start with randomly initialized policies and learn by playing with each other. Figure \ref{harvest_results} shows that the intuition from the coordination game replicates in these more complex environments: giving agents social preferences can help lead to coordination on payoff-dominant strategies. Both the risk and whether one or both agents have social preferences plays an important role. 

While the broad trends are replicated across all four games, Figure \ref{harvest_results} also emphasizes that the exact effect of prosociality is dependent on the details of the particular game. For example, in both Escalation games, but not in Markov Stag Hunt or Harvest, payoffs vary quite smoothly with the risk and prosociality. The reason is that in the Escalation games, there is a large set of joint policies which can be ordered (ie. coordinate for longer) and have increasing payoff and risk while in Harvest and Markov Stag Hunt the joint policies of interest are either those which always or never hunt/wait.

We also observe that single-agent prosociality can provide benefit in Harvest but little in Markov Stag Hunt. The reason is the relative basin of attraction of the two strategies for a single player. In the Harvest game, initial random policies have a good chance of stumbling upon corn in the `Old' phase, especially if the corn spends a high fraction of time in that phase. If one agent waits, its easy for the other agent to learn to wait. 

However, in the Markov Stag Hunt, the stag chases the agent closest to it and for random exploration to find good strategies the agent who is not being chased has to wander upon the stag. Thus it is more likely that, during initial training, the agent which is being chased steps on the stag by random exploration and will not be coordinated with its partner. In this case even a prosocial agent will learn to avoid the stag. If both agents are prosocial however, they are both encouraged to play $H$ to avoid their partner being gored.

These games illustrate that the basin of attraction for equilibria in a game with more than two strategies are influenced not only by the payoffs of the two equilibria, but also the payoffs of all the non-equilibrium policies. In the Harvest game the fraction of time in the young stage does not affect the payoff to either equilibrium but it does change the payoffs of `random' policies that pick up a mixture of young and old corn proportional to their lifetime. Similarly we observed that in a version of the Stag Hunt where the stag runs away from the agents then even for reasonably large gore penalties the agents never learn $H$ because during exploration they are unlikely to both step on the stag at the same time. This further underscores a point made by \cite{leibo2017multi} that the `difficulty' of discovering various strategies can affect equilibrium selection and furthermore that this may be dependent not only on the underlying game itself but on the function approximation used in our RL algorithm.

\section{Conclusion}
Game theoretic properties of multi-agent systems can hamper the ability of RL methods to converge to payoff-dominant policies even when such policies are equilibria. We have shown that prosocial preferences applied to even a single agent can increase the size of the basin of attraction in matrix games and that this extends to complex Markov games with function approximation. 

Prosociality can have benefits but it can also carry costs. First, if the game (or some part of the game) is not a Stag Hunt, then prosociality for a single agent may introduce new equilibria that are suboptimal for that agent. For example, in a social dilemma a prosocial agent may be exploited by its partner.\footnote{In such cases conditionally cooperative strategies should be considered \citep{lerer2017maintaining,de2012polynomial,peysakhovich2017consequentialist,littman2005polynomial}. We note here that recently proposed conditonally-cooperative algorithms for deep RL settings can actually be applied to the games studied here. Further exploring the connections between social dilemmas and coordination is an interesting question for future work.}  Second, in typical environments where an agent's actions only weakly influence on the payoffs of others, prosocial rewards increase the variance of observed rewards and thus may make reinforcement learning algorithms converge more slowly. Finally, while we assume that the agent's partner is sufficiently adaptive to learn $H$, it is better for the agent to play selfishly if the partner is not sufficiently adaptive.

We have focused on prosociality because it is a simple way to modify our agent by changing the reward function. In the related field of social dilemmas other work focuses on modifying agent's models of the environment to include the fact that other agents are also learning and so our actions today can lead to different policies for our partner tomorrow \citep{babes2008social,lerer2017maintaining,foerster2017learning}. A recent contribution to such `learning shaping' uses an explicit model of a partner's learning process included in the objective function of the agent \citep{foerster2017learning}. Such an approach has been shown to work in simple social dilemmas and so exploring it further, in particular in combination with simpler heuristic methods, is an attractive future direction.

Off-equilibrium payoffs are an important criterion for determining the basin of attraction of multi-agent optimization. However, an equally important criterion which we have not analyzed is function approximation. Deep RL does not work on the state space of a problem directly but rather approximates that state space using neural network techniques \citep{silver2016mastering,mnih2015human}. The architectures of these networks can affect the relative complexity and thus relative likelihood of different policies. This, in turn, can affect the basins of attraction for various equilibria \citep{leibo2017multi,tamar2016value}. Good choices of priors (ie. model architectures) have allowed AI to make great strides in fields like computer vision and prosocial important direction in multi-agent systems is further exploring good architectures for learning in these conditions.

We close with a practical question: how can we construct agents that can coordinate productively with humans? It has been shown that humans often fail to coordinate on payoff-dominant equilibria \citep{van1990tacit,camerer2003behavioral} but that artificial agents can be added to help groups coordinate better \cite{shirado2017locally}. We have shown that learning algorithms must take into account the learning behavior of other agents if they are to learn to coordinate in multi-agent environments. Understanding how to design human-machine hybrid systems in ways that properly account for human psychology \citep{crandall2017cooperating,kleiman2016coordinate,rand2014social,hauser2014cooperating,ouss2015punishment,arechar2016m,mao2017resilient} and also human learning strategies \citep{erev1998predicting,fudenberg2016recency} is an important practical question to extending this work.

\bibliographystyle{icml2016}  
\bibliography{20171129_GenStag_AAMAS_nonanon.bbl}

\section{Appendix}

\textbf{Proof of Theorem 1:}
To be able to use basin of attraction arguments, we must have that the structure of Nash equilibria of the game does not change as we change the utility functions. A key aspect of the Stag Hunt is that changing the prosociality parameters of the players does not delete the payoff-dominant equilibrium (since $H$ gives the highest payoff to both players), though if $c > m$ it may delete the payoff-dominated one.  In addition, given any $\alpha,$ $(F,H)$ and $(H,F)$ are never equilibria (since the player choosing $F$ always wants to deviate). Thus, it is sufficient to look at the basins of attraction of the two original equilibria.

Let $p$ be the probability that the other agent plays $Hunt$. For an agent with prosociality $\alpha$ to (weakly) prefer $Hunt$ over $Forage$ the following must hold: $$p h + (1-p) ((1 - \alpha) g + \alpha c) \geq p(\alpha g + (1 - \alpha) c) + (1-p) m$$

Solving for $p^*$, the minimum $p$ that leads the agent to play $Hunt$ yields $$p^* = \dfrac{(m- g) - \alpha(c - g)}{(h + m - g - c)}.$$ By the Stag Hunt inequalities, all three bracketed quantities must be strictly positive; therefore, $p^*$ decreases in $\alpha$ (i.e. the basin of attraction increases as $\alpha$ increases). Solving for $p^*=0$ determines the values $\alpha^*$ which makes $Hunt$ a weakly dominant strategy: $$\alpha^* = \frac{m-g}{c-g}.$$ By the Stag Hunt inequalities, $m-g \leq c-g$, and both are positive, therefore, $\alpha^* \in (0, 1]$.

\textbf{Proof of Corollary 1:}
Note that because the game restricted to any two strategies is a Stag Hunt, we can use the same argument as in Theorem 1 to look only at the original set of equilibria: the best possible equilibrium is not removed when we add prosociality (since it optimizes the joint reward), worse equilibria can be removed, and no other strategy pair $(i, j)$ can become an equilibrium.

Consider an $N \times N$ payoff matrix $U$ for agent 1; we assume symmetry, i.e. $U^{(2)} = U^{T}$ for simplicity. We can order the strategies such that for all $i<j$, $U_{ii} \geq U_{jj}$ without loss of generality. Then every subspace of $U$ is a stag hunt if for all $i<j$, $$U_{ii} > U_{ji} \geq U_{jj} > U_{ij}.$$ If agent 1 has prosociality $\alpha$, then its payoff matrix is $U^\alpha =  (1-\alpha)U + \alpha U^T.$ 

$U^{\alpha}$ is a linear interpolation between $U$ and $U^T$, so there exists $\alpha < 1$ for which $$U^{\alpha}_{ii} > U_{ij} \geq U^{\alpha}_{jj} > U^{\alpha}_{ji}$$ for all $i < j$. Thus for agent 1, $i$ weakly dominates $j$ for all $i<j$. If agent 1 plays 0, then 0 is a dominant strategy for agent 2 (for any $\alpha$).

\subsection{Details of Markov Game Training}
For grid world Markov games, we train policies $\pi$ modeled by a multi-layer convolutional neural network. For a given board size, the model has $\lceil \log_{2}(k) \rceil + 1$ repeated layers, each consisting of a 2D convolution with kernel size 3, followed by batch normalization and ReLU. The first layer has stride 1, while the successive layers each have stride 2, which decreases the width and height from $k$ to $\lceil k/2 \rceil$ while doubling the number of channels. 

We perform reinforcement learning via policy gradient using Reinforce \cite{williams1992simple} and RMSProp for optimization. The model is updated episodically in batches of 64 episodes, with a discount rate of $0.99$. Markov stag hunt and Harvest are trained for 200,000 episodes with episode length drawn from $\exp(250)$, while Escalation is trained for 500,000 episodes of length at most 50 (since Escalation only extends for the duration of coordination). For Markov Stag Hunt, we use an entropy regularization weight of $0.05$ \cite{mnih2016asynchronous}.


For Escalation Pong, we use the ALE environment modified for 2-player play as proposed in \cite{tampuu2017multiagent}. We modify the game so that each paddle hit gives a reward of $+1$ to each player, and when a 'point' is scored, the game ends and the player who scored the point receives a reward of $-a \times k$, where $k$ is the total number of hits and $a$ is a hyperparameter. We make the escalation level $k$ observable in the state by brightening a $20 \times 20$ patch in the center of the board by an amount proportional to $k$.

We train policies directly from pixels, using the pytorch-a3c package (\url{https://github.com/ikostrikov/pytorch-a3c}). 

Policies are trained directly from pixels via A3C \citep{mnih2016asynchronous}. Inputs are rescaled to $42 \times 42$ and normalized, and we augment the state with the difference between successive frames. We use 38 threads for A3C, and train for a total of 380,000 games (10,000 per thread), which takes 3 -- 4 hours on a 40-core machine. We use the default settings from pytorch-a3c: a discount rate of $0.99$, 20-step returns, and entropy regularization weight of $0.01$. We found a slightly lower learning rate of $3e-5$ was required for stable convergence.  

The policy is implemented as a convolutional neural network with four layers, following pytorch-a3c. Each layer uses a $3 \times 3$ kernel with stride 2, followed by ELU. The network has two heads for the actor and critic. We elide the LSTM layer used in the pytorch-a3c library, as we found it to be unnecessary.

\end{document}